\documentclass[runningheads]{llncs}
\usepackage{times}
\usepackage{epsfig}
\usepackage{graphicx}
\usepackage{amsmath}
\usepackage{amssymb}
\usepackage{array}
\usepackage{multirow}
\usepackage{subcaption}
\usepackage{color}
\usepackage{float}
\usepackage{graphicx}
\usepackage{mathtools}
\usepackage{hyperref}
\usepackage{fancyhdr}
\fancypagestyle{specialfooter}{%
  \fancyhf{}
  
  \fancyhead[L]{\scriptsize{Accepted in International Conference on Data Analytics and Learning, 2022}}
}

\begin{document}
\title{HWRCNet:Handwritten Word Recognition in JPEG Compressed Domain using CNN-BiLSTM Network}

\author{Bulla Rajesh\inst{1,2,*}\orcidID{0000-0002-5731-9755} \and
Abhishek Kumar Gupta\inst{1,\Psi} \and
Ayush Raj\inst{1,\Psi}\and
Mohammed Javed\inst{1,\Psi}\orcidID{0000-0002-3019-7401}\and
Shiv Ram Dubey\inst{1,\Psi}\orcidID{0000-0002-4532-8996}}
\authorrunning{Bulla. Rajesh et al.}
% First names are abbreviated in the running head.
% If there are more than two authors, 'et al.' is used.
%
\institute{Department of IT, IIIT Allahabad, Prayagraj, U.P, 211015, Idia \and Department of CSE, Vignan University, Guntur, A.P, 522213, India \\
\email{\inst{*}rajesh091106@gmail.com}\\
\email{\inst{\Psi}\{iit2018187,iit2018188,javed,srdubey\}@iiita.ac.in}}

\maketitle              % typeset the header of the contribution
\thispagestyle{specialfooter}

\begin{abstract}
Handwritten word recognition from document images using deep learning is an active research area in the field of Document Image Analysis and Recognition. In the present era of Big data, since more and more documents are being generated and archived in the compressed form to provide better storage and transmission efficiencies, the problem of word recognition in the respective compressed domain without decompression becomes very challenging. The traditional methods employ decompression and then apply learning algorithms over them, therefore, novel algorithms are to be designed in order to apply learning techniques directly in the compressed representations/domains. In this direction, this research paper proposes a novel HWRCNet model for handwritten word recognition directly in the compressed domain specifically focusing on JPEG format. The proposed model combines the Convolutional Neural Network (CNN) and Bi-Directional Long Short Term Memory (BiLSTM) based Recurrent Neural Network (RNN). Basically, we train the model using JPEG compressed word images and observe a very appealing performance with $89.05\%$ word recognition accuracy and  $13.37\%$ character error rate.

\end{abstract}
\section{Introduction}
\label{sec:intro}

Handwritten Word Recognition (HWR) is one of the most intriguing areas of research in the computer science domain, but due to the vast amount of differences in the writing styles of various people, it tends to be a bit difficult problem \cite{plamondon2000online,choudhary2013new,bulla2020}. 
Though satisfactory results have been found for isolated word recognition and character identification problems, the existing models are still not up to mark for robust handwritten word recognition in unconstrained environment \cite{sharma2021towards,8313708}. Moreover, this problem becomes more challenging when the system needs to perform the word recognition in compressed domain, and proved to be much more efficient in terms of storage and accuracy \cite{rajesh2021hh,javed2018review,mukhopadhyay2011image,florea2013compressed,rajesh2022fastss}.

\begin{figure}[!t]
\centerline{\includegraphics[scale=0.80]{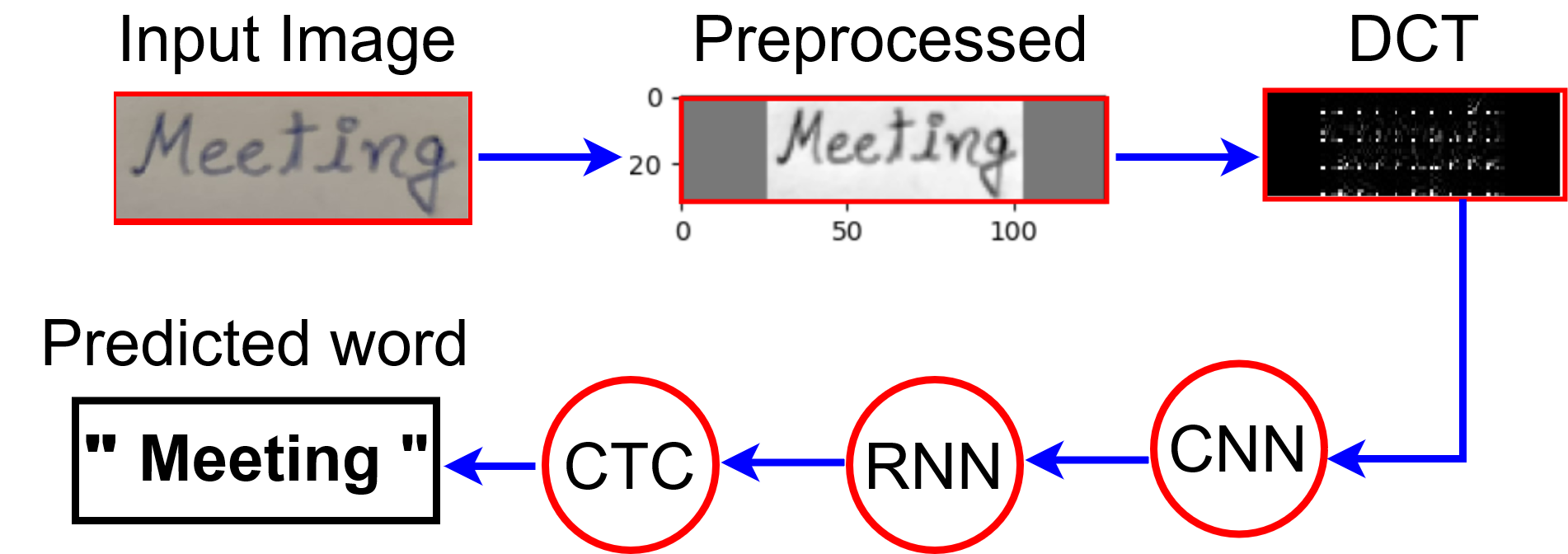}}
\caption{Overview of the proposed HWRCNet model for handwritten word recognition in compressed domain}
\label{fig:overview}

\end{figure}

General approaches in handwritten word recognition use the concept of preprocessing to reduce the vast amount of variation that are present in different ways in which the words are arranged \cite{giotis2017survey}. The common methods of preprocessing depends  upon the normalization of size of characters and correction of the slopes of various letters available in the given image \cite{b3}.
The images in the compressed domain are somewhat different to view as they do not contain the words that are generally present and can be understood by the human beings, and the features are very difficult to be extracted manually \cite{dos2020good}. So, we rely on a hybrid deep learning model consisting of a CNN and RNN. Basically, CNN extracts the important features from compressed domain images and RNN utilizes these extracted features to perform the word recognition. An overview of the proposed method is illustrated in Figure \ref{fig:overview}.

%\vspace{-10pt}
%\subsection{Literature Survey}

The handwritten character recognition from images is widely discovered problem with several approaches developed so far by utilizing the different technological solutions.
In the early days, this problem was being tackled using hand-engineered approach. Kim et al. \cite{b9} have exploited the word gaps by separating the words from a line of text image for hand-written text recognition. Later, the neural network based approaches were being utilized. España-Boquera et al. \cite{b8} have developed the Hybrid Hidden Markov Model (HMM)/Artificial Neural Network (ANN) models for improving offline handwritten text recognition in images. 

In recent years, an increasing trend has been observed for handwritten character and word recognition using deep learning approaches. In \cite{b3}, the model preciously discussed was tweaked a little by applying some modifications such as the use of a CNN-RNN model. The model used dummy  data and was utilized for achieving good efficacy by keeping some key points into consideration, such as careful initial weight assignment of the model, normalization of tilted images and exhibiting important invariances through domain specific data distortion and transformation. A hybrid approach CNN-LSTM was proposed in \cite{b4} by using CNN model along with one dimensional LSTM model. They have performed the experiments by considering the different combinations of CNN, geometric normalization, 1D LSTM, and 2D LSTM networks. They have recorded 0.25\% of test set error rate on UW3 dataset. They found that a simple 1D LSTM is outperformed by approach where a collection of various layers of convolution, 1D LSTM along with some max pooling between consecutive layers were used. One major drawback in the existing models was the inability to decode arbitrary character strings irrespective of dictionary size. In order to deal with such scenario, a Word beam search based decoding mechanism is utilized in \cite{b5}. This approach greatly affected the results as the words were constrained to be those which were already present in the set of known words. However, word beam search allowed the recognition of words which were not present in the dictionary. A LSTM-based model is also exploited in \cite{b10} scalable online handwritten text recognition system.

Through, there has been growth in technological solutions for handwritten word recognition, the focus on compressed domain has been quite limited. In \cite{b6}, a brief overview  of various methodologies was carried out for study of content based image retrieval in compressed domain on the JPEG dataset. They reported that image retrieval in compressed image domain leads to approximately $15\%$ less computational complexity as compared to image retrieval on normal images. Very recently, Rajesh et al. have developed a HH-CompWordNet model for handwritten word recognition in the compressed domain \cite{rajesh2021hh}. Their approach has directly used the CNN in the compressed domain. However, their approach does not utilize the temporal encoding capacity of RNN models, which is exploited in the proposed model in this paper. From the above context, the major contributions of this paper are given as follows- 

\begin{itemize}
    \item A hybrid HWRCNet model for handwritten word recognition in compressed domain.
    \item The model extracts the discriminative features using the CNN module and encode the temporal characteristics using the BiLSTM based RNN module.
    \item We also exploit the Connectionist Temporal Classification (CTC) loss for the training of the proposed model.
    \item Performance analysis, based on different metrics, of the proposed model on DCT JPEG compressed images.
    \item Performed extensive experiments on the proposed model on DCT JPEG compressed images to achieve state-of-the-art performance in compressed domain.
\end{itemize}

The rest of the paper is organized in 3 sections. Section 2 discuss the details of proposed architecture HWRCNet. Section 3 explains the experimental results and analysis. Finally section 4 concludes the work with a brief summary and future work.

\begin{figure}[!t]
\centerline{\includegraphics[scale=0.70]{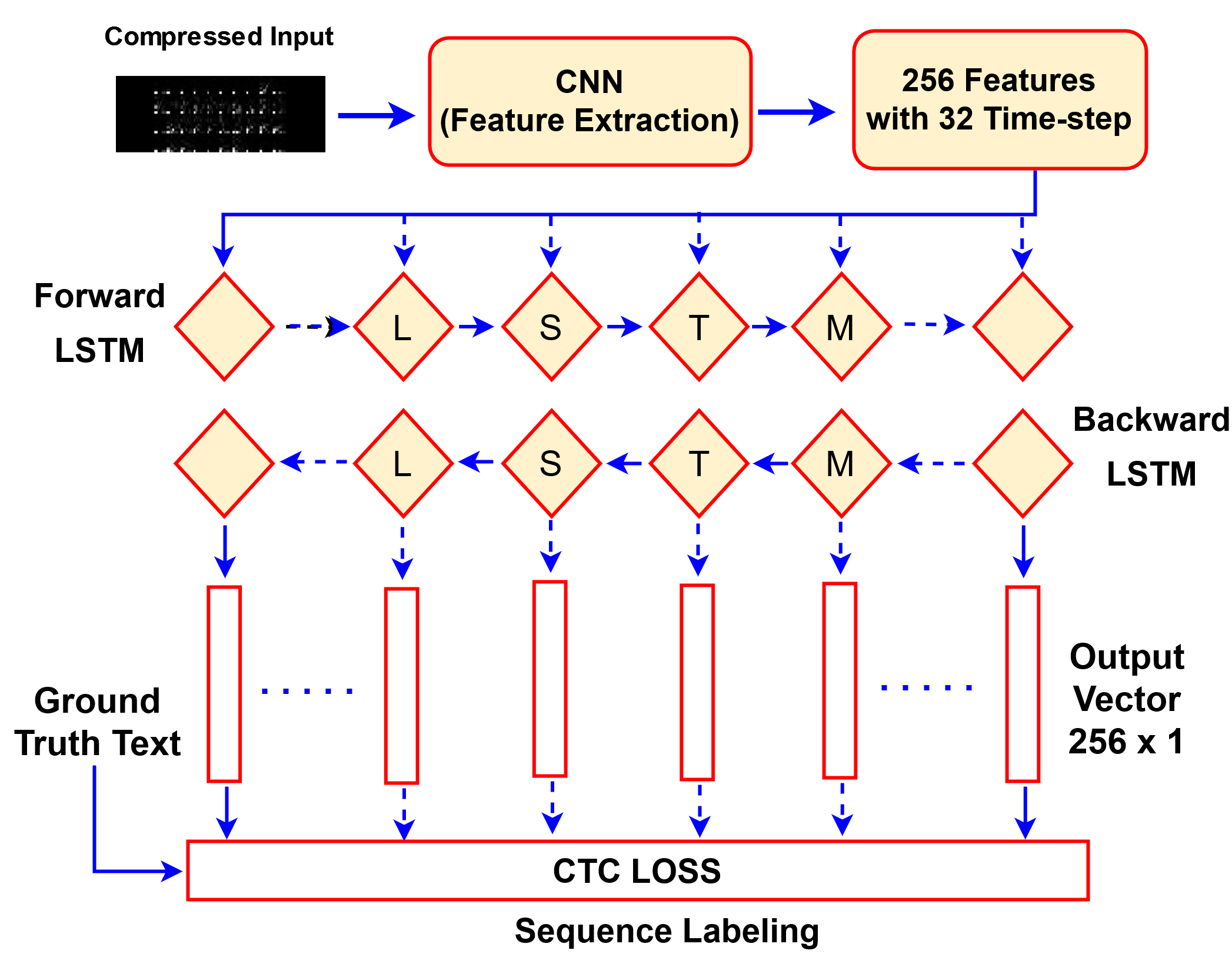}}
\caption{Schematic diagram of the proposed HWRCNet model architecture.}
\label{fig:hwrcnet}

\end{figure}
% \vspace{pt}

\section{Proposed HWRCNet Model}

\begin{figure*}[!ht]
\centerline{\includegraphics[scale=0.90]{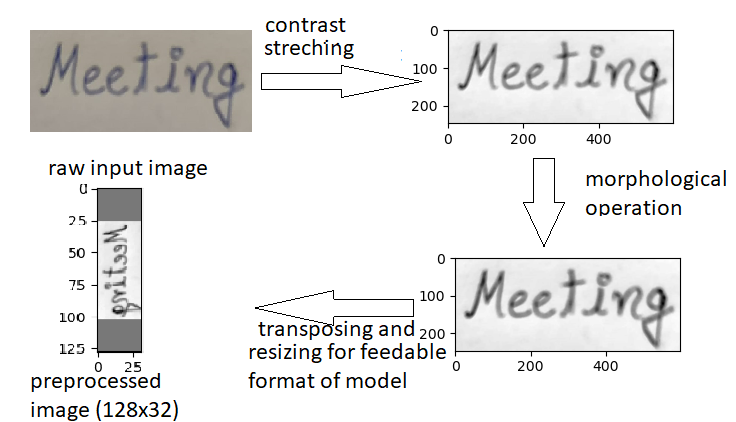}}
\caption{The pictorial visualization of all the steps involved in the preprocessing stage.}
\label{sampleimage}
\end{figure*}

\begin{figure*}[!ht]
\centerline{\includegraphics[scale=0.90]{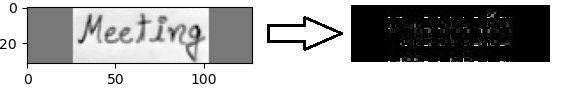}}
\caption{A sample input image in pixel domain and input stream of same document after preprocessing in compressed domain.}
\label{preprocess}
\end{figure*}

In this work, we perform the handwritten word recognition in compressed domain as shown in Figure \ref{fig:overview}. The scanned image is first converted in the compressed domain using DCT compression then word recognition is performed using the proposed hybrid CNN-BiLSTM model. We use the handwritten word recognition model \cite{scheidl2019build} as the base framework for the development of the proposed HWRCNet model as portrayed in Figure \ref{fig:hwrcnet}. It consists of following steps as discussed below. 
\textbf{Pre-Processing} - The pre-processing step as detailed in Figure \ref{fig:overview} is required to enhance the visibility of the text in the image and to ensure the required dimensionality of the input image for further processing. The overall steps involved in the preprocessing stage are shown in Figure \ref{preprocess}, and sample document image in pixel and compressed domain is shown in \ref{sampleimage}.

\textbf{Contrast Stretching}- Contrast stretching is used to enhance the contrast of the image by transforming the narrow range of intensity values into a wider range. Hence, it improves the image quality in terms of visibility as shown in the first step of Figure \ref{fig:overview}. It is given as,
\begin{equation}
{
g'(x,y)\text{ = INT}  \Bigg\{ \frac{255}{GL_{max}-GL_{min}} \Big[ g(x,y)-GL_{min} \Big] \Bigg\}    
}
\end{equation}
where $g(x,y)$ and $g'(x,y)$ are the pixel values before and after contrast stretching, respectively, $GL_{min}$ and $GL_{max}$ are minimum and maximum intensity values in the original image, respectively, and INT refers to integer.

\textbf{Resizing and Transpose} - Resizing and Transpose of the images obtained after the morphological operation is done as the last step in pre-processing to fit the images to the model. In the experiment, the resized dimension is fixed as $128 \times 32$ as shown in Figure \ref{fig:overview}. Basically, we superimpose the image with a $128 \times 32$ white patch and perform gray-level normalization to get the final pre-processed image. The resized image is further mirror transposed such that it can given as the input to tensorflow CNN as tensorflow always convolve in row-major form. After transposing, each time steps get stacked to facilitate the feature extraction of each time steps by CNN.

\subsection{DCT Compression}
The Discrete Cosine Transform (DCT) is generally used for reduction in bandwidth of image in image pre-processing. It is  also known as block compression. The block size of $8\times8$ is standard in practice. However, in our case, both $8\times8$ and $4\times4$ block size based compression is considered to observe the performance of handwritten word recognition. The DCT for a block can be given as,

\begin{equation}
DCT(i,j)=\frac{1}{\sqrt{2N}}C(i)C(j)\sum_{x=0}^{N-1}\sum_{y=0}^{N-1} f(x,y)    
\end{equation}
where $N$ is the block size, $f(x,y)$ is given as,
\begin{equation}
f(x,y) = pixel(x,y) \cos{\frac{(2x+1)i\pi}{2N}}\cos{\frac{(2y+1)i\pi}{2N}}
\end{equation}
and $C(k)$ is given as,
\begin{equation}
    C(k) = \frac{1}{\sqrt{2}} \text{ if } \: k \: \text{ is } \: 0 \text{, else } \: 1.
\end{equation}

Figure \ref{fig:overview} shows the DCT image for an example input image.

\subsection{Feature Extraction using CNN}
Convolutional Neural Networks (CNNs) can automatically learn the important visual features from images. CNNs is built with different layers such as convolution layer, max-pooling layer, activation functions, etc. The convolution layer extracts the features based on the kernel which is automatically learnt. In specific, the max-pool layer is used to detect the variation in images (i.e., detecting the brighter pixel in the DCT images). The activation functions are used to include the non-linearity in the CNN model. More specifically, CNNs are trained using the stochastic gradient descent based algorithms in order to minimize the error of the model on training dataset. In the proposed model, the 5-layer CNN receives a compressed image as input and produces a $32$ sequences of $256$ dimensional output as a feature vector which serves as the input to BiLSTM module.

\subsection{Sequence Labeling using BiLSTM}
Recurrent Neural Networks (RNNs) is the variant of neural network where the connection between various nodes follows a temporal order in order to process the temporal and sequential data. The function of RNNs can be expressed mathematically as follows:
\begin{equation}
    y_{t} = W_{hy}h_t
\end{equation}
where $y_{t}$ is output at $t^{th}$ time step, $W_{hy}$ is the parameter and $h_t$ is the hidden representation at $t^{th}$ time step and computed as,

\begin{equation}
h_{t} = f(h_{t-1},x_t) =
\text{tanh}(W_{hh}h_{t-1}+W_{xh}x_{t}) 
\end{equation}
where $x_{t}$ is the input at $t^{th}$ time step, $h_{t-1}$ is the hidden representation at ${t-1}^{th}$ time step, $W_{hh}$ and $W_{xh}$ are the parameters.

In the proposed HWRCNet model, we exploit the Bi-directional Long Short Term Memory (BiLSTM) based RNN model for the handwritten word recognition in compressed domain. Basically, the BiLSTM module received a feature vector of $32$ sequences of length $256$ from CNN as the input. The BiLSTM model performs the sequence labeling for word recognition by producing the characters for the words.
\subsection{CTC Loss}
\begin{figure}[!ht]
\centerline{\includegraphics[scale=0.20]{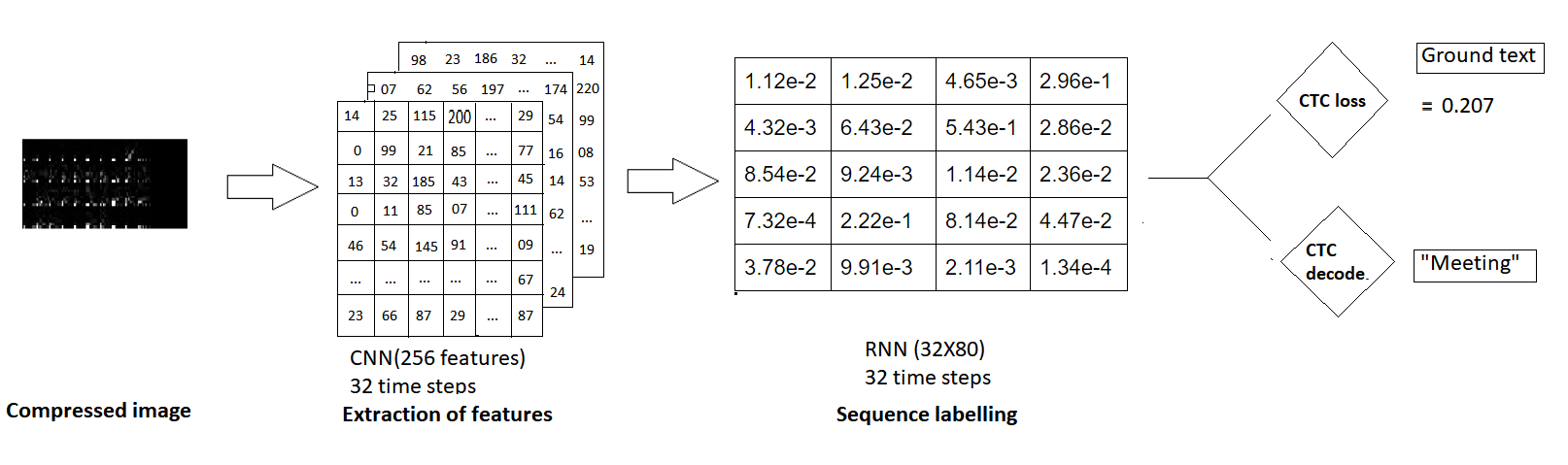}}
\caption{Feature extraction and sequence labeling.}
\label{fig7}
\end{figure}

The Connectionist temporal classification (CTC) loss is considered as the objective function for the training of the model. The CTC loss works by adding the probabilities of all probable alignments between the label and the input and defined as,
\begin{equation}
{
    p(\text{Y} | \text{X}) = \sum_{\mathrm{A} \in  \Omega_{\text{X},\text{Y}}} \prod_{\text{t=1}}^{\text{T}}  p_t(a_t | \text{X})
    }
\end{equation}
where X and Y are the input and output sequence of lengths T and U, respectively. 
Set $\Omega$ consists of all possible length sequences with length T and $a_{t}$ is the current state. The working of feature extraction using CNN, sequence labeling using BiLSTM and CTC loss computation and decoding is illustrated in Figure \ref{fig7}.

\section{Experimental Results and Analysis}
% \vspace{-3pt}
% \subsection{Dataset Description}
The benchmark IAM Handwriting Dataset \cite{b7}\cite{linkdata} is used in this paper for the experiments which comprises of images of handwritten words in English contributed by 657 people. The text accumulated to 1539 pages after being scanned, when segmented amounted to 5685 sentences which were discrete and had labels. Further segmentation broke it into 13353 lines which were discrete and had labels which further provided 115,320 words. For the purpose of breaking words and for their extraction an automatic segmentation method was used and the results were verified manually \cite{b7}.

\begin{table}[!t]
  \begin{center}
    \caption{The architecture details of the proposed HWRCNet Model.}
    \label{proposed}
    \begin{tabular}{l|l}
      \hline
      \textbf{Type} & \textbf{Description}\\ 
      \hline
        lnput  & gray-value line-image (128 x 32) \\
        Conv+Pool  & \#map 32 kernel 5 x 5, pool 2 x 2 \\
        Conv+Pool  & \#map 64 kernel 5 x 5, pool 2 x 2 \\
        Conv+Pool+BN  & \#map 128 kernel 3 x 3, pool 1 x 2 \\
        Conv+pool  & \#map 128 kernel 3 x 3,pool 1 x 2 \\
        conv+pool & \#map256 kernel 3 x 3, pool 1 x 2\\
        Collapse  & remove dimension \\
        Forward LSTM  & 256 hidden unit\\
        Backward LSTM  & 256 hidden unit\\
        Project  & project into 80 classes \\
        CTC & decode or loss \\ [1ex]
        \hline
    \end{tabular}
  \end{center}
\end{table}

The input is a grayscale image pre-processed and resized to the dimension of $128\times32$ as shown in Figure \ref{fig:overview}. We use the DCT with $8\times8$ as well as $4\times4$ block sizes for the compression of the images. The 5-layer CNN is used for feature extraction in the shape of $256$ dimensional $32$ channels which serves as sequential data for BiLSTM. The word prediction is done using BiLSTM. The model is trained using CTC loss. The dataset is divided into training and test sets with 95:5 ratio. The training is done on the compressed images in Tensorflow framework for 50 epochs with batch size of 50 having learning rate of 0.001 using ADAM optimizer with $\beta_1=0.9$, $\beta_2=0.999$, and decay $\epsilon=1e^{-08}$. We use the publicly available HTR code as our base framework \cite{linkcode}.

We use the Character Error Rate (CER) to analyze the performance of the proposed model. The CER is defined as,
\begin{equation}
{
   \text{CER = }  \frac{ \sum_{i\forall\text{samples}}^{} \text{EditDistance(} GT_i,PT_i \text{)} }{ \sum_{i\forall\text{samples}}^{} \text{\#Chars(} GT_i\text{)} } 
   }
\end{equation}
where GT is ground truth, PT is predicted text, \text{\#Chars} is the number of characters and EditDistance is a distance measure between two strings. Basically, the EditDistance function takes two strings as input and finds the minimum number of insertions, deletions or changes in one of the string to transform it into the other string, where the cost of each of the aforementioned operations is 1 and the sum of all the costs is used as the EditDistance score.
We also use the Word Error Rate (WER), Word Accuracy (WA) and Word Accuracy with Flexibility (WAF) measures to analyze the performance. 
% Basically, a word is assumed to be predicted correctly if value of EditDistance score between two words is less than 3.
The WA is the percentage of true word prediction, i.e., the EditDistance between predicted word and original word is zero. The WER is computed as 100 - WA. The WAF is the percentage of true word prediction with flexibility value two, i.e., the EditDistance between predicted word and original word is less than or equal to two.

\subsection{Results and Analysis}
%First, we perform the results comparison of the proposed HWRCNet model with CNN-RNN approach \cite{b3} on normal images for handwritten word recognition. Then we analyze the performance of the proposed model in compressed domain.
We compare the results of proposed HWRCNet model on normal images with the state-of-the-art CNN-RNN model \cite{b3}. The detailed architecture of the HWRCNet is shown in Table \ref{proposed}. Both the models are trained on the same IAM dataset. It can be noticed that the proposed model has much lower complexity and consisting of BiLSTM (i.e., Forward LSTM and Backward LSTM). The results in terms of the Word Error Rate (WER) and Character Error Rate (CER) are summarized in Table \ref{tab:normal_results}. It can be observed that the proposed HWRCNet model outperforms the CNN-RNN model in spite of having lower complexity. The main reason for the proposed model to perform better is due to the utilization of BiLSTM module which uses a sequence labeling to help in the detection of a word by using the knowledge from both forward and backward perspectives. Moreover, the usage of CTC loss function also helps to provide a better optimization of the model.
\begin{table}[!ht]
  \centering
    \caption{Results comparison for handwritten word recognition on normal images.}
    \begin{tabular}{ccc}
        \hline
        \textbf{Model} & \textbf{Word Error} & \textbf{Character Error} \\
          &  \textbf{Rate (WER)} &  \textbf{Rate (CER)} \\ \hline
        CNN-RNN \cite{b3} & 22.86 & 11.08\\
        AttentionHTR\cite{kass2022attentionhtr}& 15.40 & 6.50 \\
        \textbf{HWRCNet} 
         & 19.20 & 9.89\\
        \hline
    \end{tabular}
    \label{tab:normal_results}
\end{table}

\begin{table}[!t]
\centering
\caption{Results of the proposed HWRCNet model in compressed domain.}
\resizebox{\columnwidth}{!}{%
\begin{tabular}{c c c c c c}
\hline
\textbf{Model} & \textbf{Image Type} & \textbf{Compression Type} & \textbf{WA} & \textbf{WAF} & \textbf{CER}  \\
\hline
HWRCNet & Normal & - & 80.08 & 92.8 & 9.89 \\
HWRCNet & Compressed & $8\times8$ DCT & 69.63 & 85.76 & 14.72 \\
HWRCNet & Compressed & $4\times4$ DCT & 73.41 & 89.05 & 13.37 \\
\hline
\end{tabular}
}
\label{tab:results}
\end{table}

The results are also computed in the compressed domain using the proposed HWRCNet model and presented in Table \ref{tab:results}. The results are reported in terms of the Word Accuracy (WA), Word Accuracy with Flexibility (WAF) and Character Error Rate (CER). The results are computed under two compression settings, i.e., DCT compression using $8\times 8$ and $4\times 4$ block sizes. The results using the proposed model without compression are also provided for reference.
It can be observed that the performance of the proposed model is promising even in compressed domain also. It is also noticed that the proposed HWRCNet model is better suitable with $4\times 4$ block size based DCT compression.

\begin{figure}[!t]
\centerline{\includegraphics[scale=0.63]{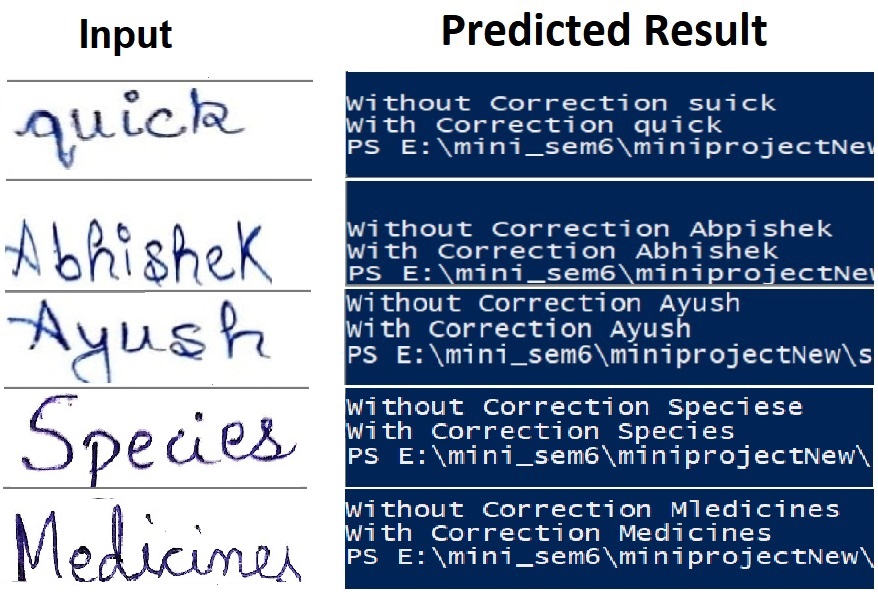}}
\caption{Few sample of the predicted results using the proposed HWRCNet method.}
\label{fig9}
\end{figure}

In order to show that the proposed model is robust enough to predict the text in random handwritten images, we perform the handwritten word recognition using our trained HWRCNet model on few images drawn by us. These results are depicted in Figure \ref{fig9}. Note that in these results without correction refers to actual predicted word and with correction refers to nearest meaningful word in dictionary w.r.t. the actual predicted word using pre-trained python library.
It can be seen that the proposed model is able to recognize the handwritten words very accurately. From the experiments it is observed that the working procedure of the proposed compressed domain approach requires $1/2(x) + y + z\%$ of time that equals $48.3\%$ and avoids $1-(1/2(x)+y+z)\%$ of time which equals $51.7\%$ of time for an uncompressed domain, as also noted by many other researchers for their research tasks \cite{gueguen2018faster,objectdetectin}. Where $x$ is the decompression time, $y$ is the operation (recognition) time, and z is recompression time.

\section{Conclusion}
In this work, we present a hybrid CNN-BiLSTM model called HWRCNet for handwritten word recognition in compressed domain. The model is directly trained using the images in the DCT compressed images. The proposed model exploits the feature extraction using CNN and sequence labeling using BiLSTM modules. The complexity of the proposed model is much lower as compared to existing CNN-RNN model. The experiments are conducted on the IAM handwritten text dataset. We observe a very promising performance of the proposed HWRCNet model in terms of the Word Accuracy and Character Error Rate. It is also noticed that the proposed model is best suitable for DCT compression with $4\times 4$ block size.
We also found that the proposed model outperforms the state-of-the-art CNN-RNN model for handwritten word recognition on normal images.

 \bibliographystyle{splncs04}
 \bibliography{ref}

\end{document}